# A Review of Developmental Interpretability in Large Language Models

Ihor Kendiukhov

Computer Science Department, Eberhard Karls University of Tübingen

Geschwister-Scholl-Platz, 72074 Tübingen

e-mail: kenduhov.ig@gmail.com

**Abstract**

This review synthesizes the nascent but critical field of developmental interpretability for Large Language Models (LLMs). We chart the field's evolution from static, post-hoc analysis of trained models to a dynamic investigation of the training process itself. We begin by surveying the foundational methodologies—including representational probing, causal tracing, and circuit analysis—that enable researchers to deconstruct the learning process. The core of this review examines the developmental arc of LLM capabilities, detailing key findings on the formation and composition of computational circuits, the biphasic nature of knowledge acquisition, the transient dynamics of learning strategies like in-context learning, and the phenomenon of emergent abilities as phase transitions in training. We explore illuminating parallels with human cognitive and linguistic development, which provide valuable conceptual frameworks for understanding LLM learning. Finally, we argue that this developmental perspective is not merely an academic exercise but a cornerstone of proactive AI safety, offering a pathway to predict, monitor, and align the processes by which models acquire their capabilities. We conclude by outlining the grand challenges facing the field, such as scalability and automation, and propose a research agenda for building more transparent, reliable, and beneficial AI systems.

**Keywords**: AI safety · AI alignment · Mechanistic interpretability · Developmental Interpretability · Large language models

**Declarations**

**Funding**. This research did not receive any specific grant from funding agencies in the public, commercial, or not-for-profit sectors.

**Conflicts of interest/Competing interests**. The authors declare that there is no conflict of interest.

# 1. Introduction: From Static Mechanisms to Developmental Trajectories

Large Language Models (LLMs) represent a paradigm shift in artificial intelligence. Their power stems from their scale, with modern models containing billions or even trillions of parameters that encode complex patterns of grammar, semantics, and world knowledge learned from their training corpora. [1]

However, this immense scale and complexity come at a cost: opacity. The internal workings of LLMs are notoriously difficult to decipher, leading them to be described as "black boxes". It is challenging to trace how a specific input leads to a particular output or to understand what abstract features the model has learned. This opacity poses significant and practical barriers to deployment, particularly in high-stakes domains such as healthcare, finance, and law. Without a clear understanding of a model's decision-making process, it is difficult to build trust, ensure reliability, diagnose failures, mitigate harmful biases, or guarantee safety. [2]

In response to this challenge, the field of Explainable AI (XAI) and, more specifically, LLM interpretability has emerged with the goal of "peering inside the 'black box'". This research endeavor seeks to uncover the mechanisms, representations, and computational pathways that give rise to the remarkable linguistic and reasoning capabilities of these models.[5] The importance of this work extends far beyond technical curiosity. It is a fundamental prerequisite for ensuring the safety, reliability, and ethical deployment of AI systems, forming a cornerstone of responsible AI development. [3]

This review focuses on a crucial and rapidly advancing subfield: developmental interpretability. While traditional *mechanistic interpretability* analyzes a fully trained, static model to reverse-engineer its learned algorithms—a process analogous to studying the anatomy of a mature organism—developmental interpretability studies how these mechanisms *form, evolve, and stabilize* during the training process. It shifts the object of study from the final artifact to the phases and phase transitions of its creation, a perspective akin to developmental biology or embryology [2]. It asks not just "How does this circuit work?" but "How did this circuit come to be, and what other circuits did it compete with and replace along the way?".

This paper posits that understanding *how* capabilities are formed is a critical, and perhaps more fundamental, challenge than understanding *what* those capabilities are in a finished model. A developmental perspective is indispensable for building robustly safe and aligned AI systems because it offers the potential for proactive intervention rather than reactive patching [4]. The difference between static and developmental analysis represents a fundamental shift in the philosophy of AI safety. Static, post-hoc analysis is inherently reactive; it can only diagnose problems or vulnerabilities in a model that has already been trained, much like an autopsy reveals a cause of death after the fact. Developmental analysis, by contrast, is proactive. By seeking to understand the "ontogeny" of model behavior, it creates the potential to monitor the training process for the precursors to unsafe cognition and to intervene directly in the learning dynamics to prevent undesirable outcomes from forming in the first place. This moves the locus of safety intervention from post-training auditing to during-training governance, which is a much more powerful and fundamental approach to alignment.

To navigate this emerging landscape, this review will proceed as follows. Section 2 outlines the foundational methodologies, from correlational probing to causal circuit analysis, that form the technical toolkit of the field. Section 3 synthesises the key initial findings on the developmental arc of LLMs, tracing the emergence of circuits, the dynamics of knowledge acquisition, and the transient nature of learned strategies. Section 4 describes in more detail the ongoing research in the developmental interpretability, including mechanistic approaches, capability emergence, cross-architectural analysis, and methodological breakthroughs. Section 5 explores powerful cognitive parallels that provide a top-down lens for interpreting these bottom-up findings. Section 6 connects these technical discoveries to their profound implications for AI safety and alignment. Finally, Section 7 outlines the grand challenges and future directions for this critical research agenda.

To provide a clear conceptual map at the outset, Table 1 delineates the different but related approaches within the broader field of LLM interpretability.

**Table 1: A Taxonomy of Interpretability Approaches**

| Approach | Primary Goal | Object of Study | Key Methodologies | Core Research Question |
|---|---|---|---|---|
| **Behavioral Analysis** | Characterize input-output behavior | Model as a black box | Adversarial testing, systematic evaluation, comparative analysis [5] | What can the model do and where does it fail? |
| **Mechanistic Interpretability** | Reverse-engineer learned algorithms | Final model weights and activations | Circuit analysis, feature visualization, causal interventions [6] | How does the model compute its output? |
| **Developmental Interpretability** | Understand the formation of capabilities | Training dynamics, model checkpoints over time | Probing across training, tracking circuit evolution, analyzing phase transitions [7] | How do learned algorithms emerge and change during training? |

## 2. Foundational Methodologies for Deconstructing Transformers

The pursuit of developmental interpretability relies on a sophisticated toolkit of analytical techniques. The evolution of these methods itself tells a story of the field's maturation, reflecting a deliberate progression from observing correlations to establishing causal mechanisms.

## 2.1. Theoretical foundations

The theoretical foundation of developmental interpretability was established through revolutionary work on mechanistic interpretability by Chris Olah and colleagues at Anthropic [8]. Their computer program analogy framework conceptualizes neural networks as compiled programs running on exotic virtual machines, where parameters function as binary instructions and activations as variables. This framework enabled systematic reverse-engineering of transformer computations into interpretable circuits.

The mathematical framework for transformer circuits introduced provided the first rigorous decomposition of transformer computations [9]. Their work identified attention heads as composed of Query-Key (QK) and Output-Value (OV) circuits, enabling path expansion methodology for analyzing information flow through transformer layers. Crucially, they discovered induction heads - circuits implementing pattern completion that became the first concrete example of interpretable computational mechanisms.

## 2.2. Representational Probing and Diagnostic Classifiers

One of the earliest and most intuitive approaches to understanding a model's internal state is representational probing. The core idea is simple: a secondary, simple classifier—the "probe"—is trained to predict a specific linguistic property, such as part-of-speech tags, syntactic dependencies, or semantic roles, using the internal activation vectors from a layer of the primary LLM as its input [10]. If the probe can predict the property with high accuracy, it is inferred that the LLM's representations encode that information.

The promise of this method lies in its ability to offer a more granular evaluation than monolithic downstream task performance. It allows researchers to ask targeted questions about what specific kinds of knowledge are present in a model's hidden states and how this information is distributed across its layers [10]. Because probes can be applied at any layer and at any point during training (using saved model checkpoints), they are a natural candidate for developmental analysis, for instance, to track how cross-lingual alignment emerges over the course of pre-training [11].

However, the probing paradigm is beset by significant methodological challenges that complicate the interpretation of its results [10]. The most fundamental issue is that of correlation versus causation. A successful probe demonstrates that a linguistic property is *correlated* with the model's activations, but it does not prove that the LLM *uses* this information in a causally meaningful way to perform its primary next-token prediction task [10]. The information might be epiphenomenal—present, but not functional.

A second major confounder is the capacity of the probe itself. A powerful, non-linear probe might be "finding" the property by performing complex computations on the LLM's representations, or worse, it could be memorizing the probing dataset's labels [10]. In such cases, the probe's success reflects its own learning ability more than the structure of the LLM's representations. This ambiguity led to a crisis of confidence in early probing studies and spurred the development of more rigorous controls, such as comparing probe performance against strong baselines and using "control tasks" to measure the probe's selectivity and ensure it is not simply learning the task from scratch [10].

### 2.3. Causal Tracing and Attribution Methods

To move beyond the correlational limits of probing, researchers developed a suite of intervention-based methods designed to establish causality. These techniques are not observational; they actively manipulate the model's internal state during a computation to measure the functional effect of specific components. The guiding question shifts from "Is information X present?" to "Does component Y causally contribute to behavior Z?" [12].

The most prominent of these techniques is activation patching, also known as causal mediation analysis. This method requires at least two inputs: a "clean" input on which the model behaves correctly (e.g., "The Eiffel Tower is in Paris") and a "corrupted" input on which it fails (e.g., "The Colosseum is in Paris"). The model is first run on the clean input, and the internal activations at a specific component (e.g., a single attention head or an MLP layer neuron) are cached. Then, during a forward pass on the corrupted input, the original activation at that component is overwritten—or "patched"—with the cached activation from the clean run. If patching this single component's state is sufficient to "rescue" the corrupted run and flip the model's output to the correct answer (e.g., from "Rome" to "Paris"), that component is deemed to have a causal role in the computation. A complementary technique, resample ablation, involves patching corrupted or zeroed activations into a clean run to see if performance breaks, thereby testing for a component's necessity. [13]

A powerful application of this approach is Causal Tracing, which was used to locate the mechanisms of factual recall in GPT models [12]. This method systematically patches states from a clean run into a corrupted run across all layers and token positions to create a "causal trace" map, identifying the critical pathway of states responsible for retrieving a fact. This work famously pinpointed specific MLP blocks in the middle layers of the transformer as the primary site of factual knowledge storage and retrieval, an insight that enabled direct model editing [12]. While other attribution methods like Integrated Gradients or LRP [14] can assign importance scores to model components, the direct, interventional nature of patching is widely considered a more rigorous and reliable test of a component's true functional role

### 2.4. Mechanistic Circuit Analysis

The ultimate ambition of mechanistic interpretability is not just to identify causally important components, but to understand the *algorithm* they collectively implement. This goal, often termed "reverse engineering," aims to deconstruct a sub-part of the neural network into a human-understandable computational graph, or "circuit".

Circuit discovery is the process of identifying these specific pathways of attention heads and MLP neurons that are responsible for a given, well-defined behavior. It typically combines causal methods like activation patching with meticulous, manual analysis of model weights and activation patterns. Through this process, researchers have successfully identified and described circuits for tasks like indirect object identification, detecting repeated sequences, and forming "induction heads"—a key circuit for in-context learning. [15]

A significant obstacle to circuit analysis is polysemanticity, where a single neuron participates in representing multiple, unrelated concepts, making its function difficult to interpret. To address this, researchers have developed techniques like Sparse Autoencoders (SAEs). An SAE is an auxiliary neural network trained to reconstruct an LLM's internal activations (e.g., from an MLP layer) via a much wider, but sparsely activated, hidden layer.[17] The goal is to decompose the dense, polysemantic activation vector into a sparse combination of more interpretable, "monosemantic" features. These learned features can be thought of as the vocabulary of the model's internal language, and identifying them is a crucial step toward describing the circuits that operate on them. [16]

The process of circuit discovery remains labor-intensive. To address this, newer methods like circuit probing aim to automate parts of the process. Circuit probing introduces a trainable binary mask over the model's weights, which is optimized to find the sparsest possible subnetwork (the circuit) that is sufficient to compute a hypothesized intermediate variable. This allows for more targeted, automated, and causal analysis of the algorithms learned by the model. [17]

Overall, the research community has moved from asking "Is property X encoded?" (a correlational question answerable by probing), to "Does component Y cause behavior Z?" (a causal question answerable by patching), to "What precise algorithm connects the input to the behavior?" (a mechanistic question answerable by circuit analysis).

## 3. Pre-History of the Developmental Interpretability: Emergence, Formation, and Transience

Equipped with the methodologies for deconstructing transformers, researchers have begun to map the developmental journey of an LLM, moving from analyzing the final product to observing the creation process. This has revealed that capabilities are not acquired in a simple, linear fashion. Instead, the training process is a dynamic and complex sequence of formation, competition, and phase transitions, where the model learns not just facts and patterns, but the very algorithms it uses to process them.

### 3.1. The Emergence and Formation of Computational Circuits

The principle that complex capabilities arise from the composition of simpler ones is fundamental to computation, and it appears to hold true for the development of transformers. Early mechanistic interpretability work on small, attention-only models provided a clear picture of this developmental hierarchy [15]. A zero-layer transformer (with only an embedding and unembedding layer) can do no more than model simple bigram statistics from the training data. A one-layer attention-only transformer learns a more sophisticated algorithm: it can function as an ensemble of bigram models and "skip-trigram" models, which can detect patterns like "A... B C" where B follows A somewhere earlier in the context.

The crucial leap occurs with two-layer models. These models can *compose* the outputs of attention heads from the first layer as inputs to the second, allowing them to implement far more complex and abstract algorithms. A canonical example of such a compositional circuit is the induction head. An induction head is a powerful, general-purpose algorithm for in-context learning that is formed by the interaction of at least two attention heads. A "previous token" head finds earlier occurrences of the current token in the context, and a second "induction" head then attends to the token that *followed* that previous occurrence to make its prediction. For example, in the sequence "[A] ->... [A] ->?", an induction head will predict. The formation of these circuits during training is a critical developmental milestone, representing the model's discovery of a generalizable copy-paste mechanism that underpins much of its in-context learning ability [15]. Tracking the precise training steps where such circuits form and stabilize is a core task of developmental interpretability [17].

### 3.2. Dynamics of Knowledge Acquisition

While early work focused on algorithmic circuits, more recent research has begun to apply a developmental lens to how models acquire and store factual knowledge. This moves beyond simply locating where facts are stored in a trained model to observing the dynamic process of their formation in what are termed "knowledge circuits". [18]

This research has revealed that the evolution of knowledge circuits is not a smooth, linear process. Instead, it follows a distinct biphasic pattern, characterized by a phase shift from formation to optimization [18].

- Formation Phase: When first encountering new knowledge during continual pre-training, the model rapidly forms a rough, structurally unstable, and computationally inefficient circuit to represent it. This phase is marked by a high degree of change in the circuit's topology (which nodes and edges are involved) and a rapid decrease in measures of circuit entropy as the model converges on a basic structure.
- Optimization Phase: After a "phase shift point," the circuit's core topology stabilizes. The subsequent training phase is dedicated to optimizing the computations within this stable structure, making knowledge representation and retrieval more efficient. This phase is characterized by a much slower rate of structural change and a focus on refining the weights of the established circuit.

This temporal evolution is mirrored by a spatial pattern of development across the model's layers, described as a "deep-to-shallow" pattern. The function of extracting knowledge from a subject token appears to develop first in the mid-to-deeper layers of the transformer. Only later, during the optimization phase, do the lower layers begin to enrich their own representations of the knowledge. This suggests a hierarchical process where higher-level layers first learn the "how" of knowledge retrieval, and lower-level layers then learn the "what." Furthermore, this entire process is scaffolded by prior knowledge; new facts that are relevant to concepts the model already understands are integrated into circuits much more efficiently than completely novel information. [18]

### 3.3. Phase Transitions and the Nature of Emergent Abilities

Perhaps the most debated and intriguing aspect of LLM development is the phenomenon of emergent abilities. These are capabilities, such as multi-step arithmetic or understanding nuanced humor, that appear to be absent in smaller-scale models but suddenly manifest in larger models [19]. This non-linear improvement, where quantitative changes in scale (e.g., parameters, training data, compute) lead to qualitative changes in behavior, is a hallmark of complex systems and has been a central topic of discussion in AI. [20]

The discovery of phase changes during training by Olsson et al. (2022) fundamentally transformed understanding of LLM development. Their analysis of 34 transformer models revealed that all models larger than one layer undergo an abrupt transition during training around 2.5-5 billion tokens. Before this phase change, models demonstrate less than 0.15 nats of in-context learning capability. After the transition, they achieve approximately 0.4 nats consistently across model sizes. [21]

This phase change coincides precisely with the formation of induction heads through a process called K-composition and Q-composition. Previous token heads in early layers develop first, followed by induction heads in later layers that compose with these earlier mechanisms. The developmental timeline shows sequential formation: simple circuits form before complex ones, with progressive complexity emerging through training.

The existence of emergence itself has been contentious. Some researchers have argued that it is an illusion, a "mirage" created by the choice of non-linear metrics (like accuracy) that only show a signal after a certain performance threshold is crossed. They contend that with more sensitive, continuous metrics, performance scales smoothly and predictably [22]. However, other studies have directly tested this claim and found that sharp, emergent-like jumps in performance persist even when using continuous metrics, suggesting that emergence is a real feature of the learning dynamics, not just a measurement artifact [23].

A crucial insight that helps resolve this debate reframes the discussion away from model scale and towards the training process itself. Research indicates that a model's pre-training loss—a direct measure of how well it predicts the training data—is a more fundamental predictor of downstream capabilities than abstract scaling variables like parameter count [20]. From this perspective, emergent abilities can be understood as phase transitions that occur during training [24]. A model's performance on a difficult, emergent task will remain at random chance until its overall pre-training loss crosses a specific critical threshold. Below this loss value, the model appears to have learned a sufficiently powerful and general set of internal representations and algorithms to begin solving the new task, after which its performance on that task improves steadily [23]. This framework provides a concrete, measurable mechanism for the abstract concept of emergence, linking it directly to the developmental trajectory of the model's learning process.

### 3.4. The Transient Nature of Learned Strategies

The developmental story is further complicated by the discovery that the algorithms a model learns are not necessarily permanent. Models have access to multiple potential strategies for solving a given task, and the one that appears first may not be the one that ultimately survives the optimization process.

A key example of this is the dynamic between In-Context Learning (ICL) and In-Weights Learning (IWL) [7]. ICL is the ability to adapt behavior at inference time based on examples provided in the prompt, a flexible and general strategy. IWL involves encoding knowledge or a task solution directly into the model's parameters (weights), a more specialized and rigid strategy. Groundbreaking research using synthetic datasets has shown that ICL is often a transient phenomenon [7]. When trained on data where both strategies are viable, transformers often first develop ICL circuits to solve the task. This is an intuitive result, as ICL is a more general-purpose solution. However, if training continues long enough ("overtraining"), the model may discover a more efficient IWL solution. In a striking display of dynamic self-optimization, the model will then dismantle the ICL circuit and replace it with the IWL circuit, even as the overall training loss continues to decrease.

This suggests an ongoing competition between circuits for the finite representational capacity within the model. The learning process, driven by gradient descent, does not simply add new capabilities; it actively restructures and replaces existing ones in a relentless search for more optimal solutions to the global prediction objective. This finding has profound implications, suggesting that a model's apparent reasoning strategy is not a fixed property but a developmental stage that could change with further training. Some evidence suggests that techniques like L2 regularization may help stabilize ICL and make it a more persistent strategy [7].

These distinct developmental phenomena—the compositional formation of circuits, the biphasic acquisition of knowledge, emergent phase transitions, and the transience of learning strategies—are not isolated events. They are different windows into the same fundamental underlying process: a high-dimensional search through the space of possible algorithms, guided by the simple objective of minimizing prediction error. The model is not merely a passive recipient of information; it is a dynamic system that actively learns *how to learn*. The "deep-to-shallow" knowledge pattern and the "ICL-to-IWL" strategic shift are observable traces of this algorithmic self-improvement. The developmental arc of an LLM is the story of this search, as the model builds, refines, and replaces its own internal machinery to become an ever-more-efficient prediction engine.

### 3.5. Grokking Phenomena

Grokking phenomena discovered by Power et al. (2022) revealed that neural networks can achieve perfect generalization long after apparent overfitting [25]. This delayed generalization involves a transition from memorization to algorithmic understanding, with sudden jumps from near-zero to perfect performance occurring thousands of epochs after training loss convergence. Recent work has demonstrated grokking during large-scale LLM pretraining, showing that different capabilities enter grokking phases asynchronously.

Table 2 provides a concise summary of these key developmental findings.

**Table 2: Summary of Key Early Developmental Findings in Transformers**

| Phenomenon | Description | Key Mechanism/Driver | Model Context |
|---|---|---|---|
| **Circuit Composition** | Simple circuits like skip-trigrams compose to form complex algorithms like induction heads. | Composition of attention heads across layers. | 1-2 Layer Attention-Only Transformers |
| **Knowledge Acquisition** | Biphasic evolution from inefficient 'formation' to stable 'optimization' of knowledge circuits. | Deep-to-shallow pattern of function development. | Continual pre-training on GPT-2, Llama |
| **Emergent Abilities** | Sudden, non-linear jumps in capability on specific tasks. | Phase transition when pre-training loss crosses a critical threshold. | Scaling across model families (GPT-3, PaLM) |
| **ICL Transience** | In-context learning (ICL) emerges then disappears, replaced by in-weights learning (IWL). | Competition between ICL and IWL circuits for model capacity. | Synthetic data experiments on transformers |

## 4. Research Findings within the Developmental Interpretability Agenda

### 4.1. Mechanistic Approaches During Training Dynamics

Mechanistic analysis of training dynamics has revealed sophisticated patterns of circuit formation and evolution. Research by Nanda et al. (2023) using discrete Fourier transforms on modular arithmetic tasks revealed training splits into three continuous phases: memorization, circuit formation, and cleanup [26]. The circuit formation phase involves gradual development of generalizing circuits while maintaining memorization circuits, followed by elimination of memorization and retention of generalization circuits [27].

Attention head specialization during training follows predictable patterns across architectures and random seeds. Duan et al.'s analysis using Syntactic Specialization Index (SSI) across 46 training checkpoints revealed critical periods where rapid internal specialization occurs [28]. Syntactic sensitivity emerges gradually but concentrates in specific layers, with the process showing consistency across different model scales and training configurations.

Feature evolution analysis through sparse autoencoders (SAEs) has provided unprecedented insight into training dynamics. The SAE-Track framework by Xu et al. (2024) enables continuous tracking of feature formation, semantic evolution, and directional drift during training [29]. Features emerge through systematic processes with splitting and absorption phenomena, while their semantic meanings evolve continuously throughout training.

Layer-wise representation changes during training reveal systematic patterns of information processing evolution. Research shows that intermediate layers often outperform final layers by up to 16% on embedding tasks, with strong entropy reduction at intermediate layers tied to residual path merging. Training significantly shifts layer-wise representation quality, with different patterns emerging in autoregressive versus non-autoregressive models. [30]

Advanced methodological frameworks include activation patching and causal interventions that enable systematic testing of component importance during training. Best practices distinguish between denoising (clean → corrupt) and noising (corrupt → clean) approaches, with different applications for circuit discovery versus validation. Path patching techniques test direct versus mediated component interactions across training stages. [31,32]

Attribution-based methods have scaled mechanistic analysis to industrial applications. Attribution patching uses gradient-based approximations to compute patching effects for all components in a single backward pass, while Layer-Wise Relevance Propagation (LRP) provides attention-aware attribution covering both input and latent representations. [33]

## 4.2. Capability Emergence Patterns and Learning Phases

Emergent abilities in LLMs are characterized by capabilities absent in smaller models that appear suddenly in larger models. These abilities exhibit sharp transitions rather than gradual improvements, with performance remaining near-random until critical thresholds, then jumping to above-random performance [19]. Wei et al. (2022) documented over 100 emergent abilities with specific scale thresholds: few-shot arithmetic emerges at ~$2.3 \times 10^{22}$ FLOPs (13B parameters),

MMLU benchmark capabilities at ~$3.1 \times 10^{23}$ FLOPs (175B parameters), and complex reasoning tasks at even higher scales [34].

Learning curves and developmental trajectories exhibit three distinct phases analogous to human language development. Chang and Bergen's (2022) analysis reveals a babbling phase with random token generation and high variability, followed by syntax acquisition with grammatical structure learning and increased output diversity, culminating in semantics acquisition with meaning representation development and coherent generation [35].

Phase transitions in neural networks exhibit characteristics analogous to physical systems, with sharp transitions at critical parameter values, order parameters quantifying system state changes, and correlation lengths measuring transition sharpness. Research reveals triple phase transitions during training: initial brain alignment and instruction following, temporary brain detachment and stagnation, and final brain realignment with capability consolidation [36].

Critical periods and sensitive phases mirror human language development, with early training crucial for basic linguistic patterns, mid-training critical for syntactic rule acquisition, and late training necessary for semantic and pragmatic understanding. Models show decreased ability to acquire new capabilities as training progresses, similar to biological neural plasticity reduction [37].

Sharp versus gradual capability transitions represent fundamentally different types of emergence. Sharp transitions include few-shot arithmetic with sudden activation at 13B parameters and chain-of-thought reasoning emerging at 100B parameters. Gradual transitions include perplexity improvements following predictable scaling laws and basic language modeling showing continuous enhancement [38].

Quantitative studies have established methodological frameworks including scaling law analysis with power law fitting to identify emergence points, cross-entropy analysis revealing hidden improvements before apparent emergence, and probing studies measuring internal representation quality and layer-wise capability development [19,39,40].

**4.3. Recent Methodological Advances and Breakthrough Discoveries**

Sparse autoencoder scaling breakthroughs in 2024-2025 have revolutionized developmental analysis capabilities. OpenAI's k-sparse autoencoder methodology successfully trained a 16 million feature autoencoder on GPT-4 activations using 40 billion tokens, discovering clean scaling laws with respect to autoencoder size and sparsity. This breakthrough eliminates complex hyperparameter tuning while achieving unprecedented reconstruction fidelity equivalent to ~10× less compute models [41].

Attribution graph methodology by Anthropic represents a paradigm shift in circuit analysis. Cross-layer transcoders (CLTs) with 30 million features across layers enable computational flow tracing through attribution graphs revealing causal interactions between features. This methodology successfully analyzed multi-step reasoning, planning in poetry, multilingual circuits, and internal arithmetic strategies, providing satisfying insights for approximately 25% of analyzed prompts [42].

Dynamic feature evolution tracking through the SAE-Track framework enables continuous monitoring of feature formation processes, semantic evolution, and directional drift during training. This methodology provides mechanistic insights into training dynamics previously impossible to study, revealing systematic patterns in feature emergence, splitting, and absorption phenomena [43].

Information-theoretic progress measures using O-Information quantify emergent properties without supervision. Synergy and redundancy metrics predict grokking phenomena and identify emergent sub-networks, providing task-independent tools for understanding phase transitions and capability development [44].

Interactive visualization platforms have democratized access to interpretability insights. OpenAI's SAE Viewer provides public access to GPT-2 and GPT-4 features with real-time activation analysis, while Anthropic's Attribution Graph Interface enables complex exploration of computational flows with feature grouping and validation through intervention experiments.

Advanced evaluation metrics now include feature quality assessment through recovery of hypothesized features, explainability of activation patterns, and sparsity of downstream effects. All metrics show consistent improvement with autoencoder size, enabling systematic comparison of interpretability approaches across different scales and architectures.

A summary of recent advances in developmental interpretability is given in Table 3.

**Table 3: Summary of Recent Developmental Findings in Transformers.**

| Category | Key Research Area | Core Findings & Discoveries | Illustrative Examples & Methods |
|---|---|---|---|
| **Mechanistic Approaches** | Circuit & Feature Evolution | Training follows distinct phases (memorization, formation, cleanup); features evolve systematically through splitting and absorption. | Discrete Fourier transforms, Sparse Autoencoders (SAE-Track), attribution patching. |
| | Attention & Layer Dynamics | Attention heads specialize during critical periods; intermediate layers often outperform final layers in representation quality. | Syntactic Specialization Index (SSI), activation patching, Layer-Wise Relevance Propagation (LRP). |
| **Capability Emergence** | Emergent Abilities & | Capabilities can appear suddenly at specific model scales rather | Few-shot arithmetic emerges at ~13B parameters; MMLU at ~175B parameters. |

| | Scaling | than improving gradually. | |
| --- | --- | --- | --- |
| | Learning Phases & Transitions | LLM learning mirrors human development (babbling, syntax, semantics) and exhibits sharp phase transitions like physical systems. | Scaling law analysis, probing studies, cross-entropy analysis. |
| **Cross-Architectural Analysis** | Architectural Impact | Different architectures (e.g., BERT, GPT, T5) produce distinct developmental patterns and scaling efficiencies. | GPT's causal attention leads to sharp emergence; BERT's bidirectional attention gives steady improvement. |
| | Scaling & Efficiency | Performance follows predictable power laws (Chinchilla Laws); architectural choices (RoPE, SwiGLU) significantly improve efficiency. | LLaMA-13B outperforms GPT-3-175B due to superior architecture and training. |
| **Recent Breakthroughs** | Methodological Advances | Scalable sparse autoencoders and attribution graphs have revolutionized circuit and feature analysis. | OpenAI's 16M feature k-sparse AE; Anthropic's Attribution Graphs with cross-layer transcoders (CLTs). |
| | New Frameworks & Tools | New methods track feature evolution continuously, quantify emergence without supervision, and democratize insights. | SAE-Track for dynamic feature monitoring, O-Information for progress measures, interactive SAE viewers. |

## 5. Cognitive Parallels: LLM Development Through a Human Lens

While the bottom-up, mechanistic investigation of circuits and activations provides a granular view of LLM development, it can be difficult to see the forest for the trees. To complement this approach, researchers have increasingly turned to top-down frameworks from cognitive science and linguistics. Drawing analogies to human cognitive and linguistic development provides powerful, high-level hypotheses about the nature of LLM learning, helping to organize and interpret the complex findings from circuit-level analysis.

### 5.1. Simulating the Trajectory of Linguistic Development

At a high level, the learning trajectory of an LLM shows striking parallels to that of a human child. Studies that prompted LLMs to adopt personas of children at different ages found that the models' generated language systematically increased in complexity with the simulated age [45]. This was measured both by simple metrics like utterance length and more sophisticated ones like Kolmogorov complexity. A particularly notable "growth spurt" in linguistic complexity was observed between the simulated ages of one and two, mirroring a critical period in human language acquisition [45].

This behavioral parallel may have a deeper, structural analog in the model's representations. In the developing human brain, language processing matures hierarchically: the ability to represent fast, low-level phonetic features develops early in the superior temporal gyrus, while the capacity for slower, higher-level word and concept representations emerges later in the associative cortices [46]. Remarkably, this neuro-developmental trajectory appears to be spontaneously recapitulated by LLMs during their training. Analysis of model checkpoints shows that the representations learned by LLMs evolve in a manner that aligns with this biological maturation, progressing from encoding child-like linguistic features to those identifiable only in the adult human brain [46]. This suggests that the hierarchical structure of language may impose fundamental constraints on any learning system, biological or artificial, leading to similar developmental pathways.

**5.2. Analogy vs. Abstraction in Generalization**

A central question in both cognitive science and AI is how systems generalize from finite experience to novel situations. A widespread assumption about LLMs was that, given their mastery of grammar, they must be learning abstract linguistic rules. However, a compelling body of evidence now suggests that LLMs, much like humans, generalize primarily through analogy to stored examples rather than through the application of explicit, symbolic rules [47].

In one study, when an LLM was tasked with turning a made-up adjective (e.g., "friquish") into a noun, it did not appear to consult a learned grammatical rule. Instead, its choice ("friquishness") was best explained by the word's similarity to other known words (like "selfish") that take the "-ness" suffix. The model's behavior suggests it is constantly, implicitly asking, "What does this new thing remind me of?" and basing its response on the patterns associated with the most similar items in its memory [47]. This analogical reasoning process has been identified as a cornerstone of human cognition and creativity, emerging in children around age three without explicit instruction [48]. Studies on more complex, story-based analogical reasoning tasks confirm that LLMs can extract relational similarities, though they may struggle with more distant analogies compared to humans and require careful prompting to map higher-order causal structures [49].

While the *process* of analogical reasoning appears similar, there is a crucial difference in the underlying data structure, which reveals a key cognitive divergence between humans and current LLMs. Humans are thought to acquire an abstract "mental dictionary", a store of word *types* and their properties. We recognize that "cat" and "cats" are instances of the same underlying lexical entry. LLMs, in contrast, do not appear to form this level of abstraction. Instead, they seem to treat every single instance of every word encountered during training as a unique "memory trace" [47]. They generalize not over a compact dictionary of abstract types, but directly over a colossal, instance-based memory.

This lack of abstraction may be the key to understanding one of the great puzzles of AI: why LLMs require orders of magnitude more data to learn a language than a human child. They appear to compensate for a less abstract, less data-efficient learning mechanism with a memory of examples that is unimaginably vast by human standards.

These cognitive findings provide a powerful, unifying theory that can bridge low-level mechanistic discoveries with high-level model behavior. The observation that LLMs learn via analogy over specific instances suggests that the "circuits" being painstakingly reverse-engineered are not implementations of formal linguistic rules, but are instead highly optimized, massively parallel similarity-matching engines. From this perspective, the fundamental computation of a transformer is analogical pattern matching. The developmental processes observed by interpretability researchers—the composition of attention heads into induction heads, the ICL-to-IWL shift, the biphasic optimization of knowledge circuits—can all be re-interpreted as different facets of this analogical engine being constructed and refined. For example, the shift from ICL to IWL can be seen as the model optimizing its analogical reasoning process by "caching" the results of frequently encountered analogies directly into its weights for faster retrieval. This cognitive lens suggests that an LLM's "understanding" is not rooted in abstract knowledge of rules, but in an unparalleled memory of specifics and an incredibly sophisticated mechanism for reasoning by similarity.

## 6. Implications for AI Safety and Alignment

The insights gleaned from developmental interpretability have profound and urgent implications for the field of AI safety, which seeks to ensure that advanced AI systems are robust, trustworthy, and aligned with human values. By illuminating the process by which capabilities are formed, this research offers a new paradigm for tackling the alignment problem.

### 6.1. A Paradigm Shift: From Post-Hoc Auditing to Proactive Governance

The ultimate promise of interpretability for AI safety is that understanding leads to control. If we can understand why a model produces a harmful output, we can work to fix it. Developmental interpretability pushes this promise a crucial step further: by understanding how unsafe or undesirable cognition *forms* during training, we can potentially develop interventions to prevent it from emerging in the first place. This represents a fundamental shift from a reactive safety posture, which relies on post-hoc auditing and red-teaming of a finished model, to a proactive one focused on developmental governance.

Instead of waiting for a model to demonstrate dangerous capabilities like deception, manipulation, or autonomous hacking, a developmental approach would involve actively monitoring the training process for the emergence of the underlying circuits or the crossing of the phase transition thresholds associated with such capabilities. For example, if we could identify a "deception circuit," we could monitor for its formation and potentially use targeted regularization techniques to penalize its development. This would transform safety from a process of black-box testing to one of white-box, developmental monitoring, providing a much-needed early-warning system.

### 6.2. Taming Unpredictability and Emergence

As discussed, one of the most significant risks in AI development is the unpredictable emergence of powerful new capabilities. A model that suddenly and unexpectedly gains a dangerous skill is far more difficult to manage and contain than one whose capabilities evolve along a predictable path. This unpredictability makes it challenging to anticipate risks and implement safeguards ahead of time.

Developmental interpretability offers a direct path toward taming this unpredictability. By tracking developmental trajectories and, crucially, by linking the phenomenon of emergence to measurable quantities like pre-training loss, this research provides a framework for making capability gains more predictable. If future research can reliably identify the "loss thresholds" at which certain classes of capabilities (e.g., advanced reasoning, strategic planning) tend to emerge, then we can anticipate when they are likely to appear during a training run. This would allow safety researchers and developers to prepare for and evaluate these new capabilities in a controlled manner, transforming "unknown unknowns" into "known unknowns" and significantly reducing strategic surprise.

**6.3. Addressing the Core Alignment Problem**

The AI alignment problem is the grand challenge of ensuring that the goals and behaviors of highly capable AI systems are robustly and reliably aligned with human values and intentions The core principles of alignment are often summarized as robustness (reliability under adverse conditions), interpretability (understandable decision-making), controllability (responsiveness to human intervention), and ethicality (adherence to moral principles).

Developmental interpretability reframes the alignment problem in a deeper and more fundamental way. The goal is not merely to align the final *outputs* of an opaque black box, for instance by fine-tuning it on examples of desired behavior. Instead, the goal becomes to align the *learning process* itself, ensuring that the internal representations, algorithms, and world-models the model develops are themselves structured in a way that is conducive to alignment.

This distinction is of paramount practical importance. Consider the finding that in-context learning (ICL) can be a transient developmental phase [7]. This poses a significant and subtle threat to the stability of our current leading alignment techniques, such as Reinforcement Learning from Human Feedback (RLHF) and Constitutional AI. These methods work by providing the model with examples and rules in-context, effectively leveraging the model's ICL capabilities to steer its behavior towards being more helpful and harmless. However, the transience of ICL implies that this learned alignment might be built on a temporary foundation. With further training or optimization, a model might replace the *general principle* of harmlessness (an ICL strategy) with a set of *specific, memorized rules* about what not to say in certain situations (an IWL strategy). This IWL-based alignment would likely be far more brittle, less generalizable, and more susceptible to novel "jailbreak" attacks that fall outside the narrow distribution of its alignment fine-tuning data. It has learned to mimic alignment in familiar contexts, but has lost the underlying reasoning circuit for it. Therefore, developmental interpretability is not just an optional add-on for safety; it is an essential tool for validating the long-term stability and robustness of our primary alignment techniques. We must use it to ensure we are instilling deep, generalizable principles, not just shallow, brittle behaviors.

**6.4. Current limitations and methodological challenges**

Coverage limitations remain significant across current methodologies. Attribution graphs provide satisfying insights for only ~25% of analyzed prompts, SAEs capture only fractions of total model computation, and attention circuit mechanisms remain poorly understood. Even with millions of features, complete model coverage requires potentially billions or trillions of features for frontier models.

Feature ontology challenges include poorly defined concepts of what constitutes a "feature," complications from feature splitting and merging during interpretation, and unclear relationships between discovered features and actual computational processes. Feature universality across models and scales remains an open question requiring systematic investigation [50].

Validation difficulties include intervention experiments requiring unnatural activation strengths, risk of incomplete or misleading circuit explanations, and limited ability to verify explanations at scale. The challenge of establishing true causal relationships in high-dimensional spaces with potential spurious patterns requires more rigorous validation frameworks [51].

Computational resource requirements create barriers to widespread adoption. Training 16M feature SAEs required massive computational resources with current methods requiring 10× compute loss for full model reconstruction. Scaling to complete frontier model analysis may require unprecedented computational investments.

# 7. Synthesis, Open Problems, and Future Directions

## 7.1. Synthesis of Findings

The field of developmental interpretability is painting a new and dynamic picture of how Large Language Models operate. It reveals that they are not static artifacts but complex systems that undergo a rich, multi-stage "ontogeny" during training. We now understand that LLMs learn by composing simple computational building blocks into sophisticated, hierarchical algorithms, such as the induction heads that enable in-context learning. The acquisition of new knowledge and capabilities is not linear but is marked by distinct phase transitions, often linked to measurable quantities like pre-training loss, where new abilities emerge and different learning strategies compete for dominance and are sometimes discarded. This developmental process, while alien in its computational substrate and scale, shows intriguing and informative parallels with human cognitive development, particularly in its fundamental reliance on analogical reasoning over vast stores of remembered examples. Ultimately, this review has argued that charting this developmental arc is not an esoteric academic exercise but a critical frontier for AI safety, offering a new, proactive paradigm for predicting, monitoring, and aligning the powerful AI systems of the future.

**7.2. Grand Challenges and Open Problems**

Despite rapid progress, the field faces several formidable challenges that must be overcome to realize its full potential.

- **Scalability:** This is arguably the most significant barrier. The most detailed, circuit-level analyses have been performed on small, often "toy" models with only a few layers. Applying these meticulous, manual techniques to frontier models with hundreds of billions or trillions of parameters is a monumental task. The complexity of potential circuits grows exponentially, making current methods computationally and cognitively intractable at scale.
- **Automation:** The process of identifying features and reverse-engineering circuits is currently a painstaking, artisanal effort that relies heavily on expert intuition and laborious experimentation. For interpretability to keep pace with the development of new models, the automation of circuit discovery and hypothesis generation is essential.
- **Universality and Completeness:** Do the circuits and features identified in one model (e.g., GPT-2) generalize to others (e.g., Llama, Claude)? Can we find a universal "vocabulary" of features and a "standard library" of circuits that are common across different architectures and training runs? Furthermore, how can we ever be sure that an explanation for a given behavior is complete, and not just a partial story that misses other crucial contributing pathways?
- **The Problem of Immense Cost and Opacity:** A practical and systemic challenge is that many state-of-the-art LLMs are proprietary and accessible only through limited APIs. This prevents the direct access to model weights and internal activations that is necessary for nearly all mechanistic and developmental interpretability techniques. Even for open-source models, their immense size makes running the necessary experiments prohibitively expensive for most academic labs.
- **Hallucinated Explanations:** A promising avenue for interpretability is to use LLMs to explain their own reasoning. However, this approach is undermined by the very problem it seeks to solve: LLMs can "hallucinate" plausible-sounding but incorrect or baseless explanations for their behavior. This makes it difficult to trust the model's self-explanation without external verification, creating a challenging circular problem.[11]

**7.3. Future Research Agenda**

Addressing these challenges points toward a clear research agenda for the coming years.

- **Developing a Unified Theory:** The field currently consists of a collection of fascinating but somewhat disparate findings. A key goal should be to move towards a more unified theoretical framework of developmental dynamics in deep learning, potentially drawing on concepts from statistical physics, dynamical systems, and developmental biology to explain phenomena like phase transitions and circuit competition.
- **Robust Benchmarks and Metrics:** The field needs standardized benchmarks and evaluation

metrics to rigorously assess the quality, faithfulness, and completeness of interpretations.[26] Just as tasks like GLUE and SuperGLUE drove progress in NLP capabilities, a suite of "Interpretability-BENCH" tasks could drive progress in understanding.

- **Interpretability Across Training Paradigms:** Developmental analysis must expand to cover the full lifecycle of a modern LLM. This includes understanding the distinct effects of pre-training, instruction fine-tuning, and reinforcement learning (RLHF), and crucially, how these different training phases interact to shape the final model's representations and algorithms.[54]
- **Expanding to New Frontiers:** The principles of developmental interpretability must be applied to the next generation of AI systems. This includes understanding the formation of capabilities in multimodal models that integrate vision and language, and, most critically, in agentic AI systems that can take actions in the real world, where the safety stakes are dramatically higher.
- **Collaborative and Human-in-the-Loop Interpretability:** Given the complexity of the task, the future of interpretability is likely collaborative. This involves developing new interfaces and tools that allow human experts and AI systems to work together on the problem of interpretation, leveraging the pattern-matching strengths of AI and the conceptual reasoning abilities of humans.

By pursuing this agenda, the field can continue to peel back the layers of opacity surrounding our most advanced AI systems, moving us closer to a future where we can not only use these powerful tools, but truly understand and trust them.

## 8. References

1. Naveed, H., Khan, A. U., Qiu, S., Saqib, M., Anwar, S., Usman, M., ... & Mian, A. (2023). A comprehensive overview of large language models. ACM Transactions on Intelligent Systems and Technology.
2. Bereska, L., & Gavves, E. (2024). Mechanistic interpretability for AI safety--a review. arXiv preprint arXiv:2404.14082.
3. Xu, F., Uszkoreit, H., Du, Y., Fan, W., Zhao, D., & Zhu, J. (2019, September). Explainable AI: A brief survey on history, research areas, approaches and challenges. In CCF international conference on natural language processing and Chinese computing (pp. 563-574). Cham: Springer International Publishing.
4. Mapping the Mind of a Large Language Model - Anthropic, accessed August 12, 2025, https://www.anthropic.com/research/mapping-mind-language-model
5. Grey, M., & Segerie, C. R. (2025). Safety by Measurement: A Systematic Literature Review of AI Safety Evaluation Methods. arXiv preprint arXiv:2505.05541.
6. Conmy, A., Mavor-Parker, A., Lynch, A., Heimersheim, S., & Garriga-Alonso, A. (2023). Towards automated circuit discovery for mechanistic interpretability. Advances in Neural Information Processing Systems, 36, 16318-16352.

7. Singh, A., Chan, S., Moskovitz, T., Grant, E., Saxe, A., & Hill, F. (2023). The transient nature of emergent in-context learning in transformers. Advances in neural information processing systems, 36, 27801-27819.
8. Olah, C., Satyanarayan, A., Johnson, I., Carter, S., Schubert, L., Ye, K., & Mordvintsev, A. (2018). The building blocks of interpretability. Distill, 3(3), e10.
9. Olsson, C., Elhage, N., Nanda, N., Joseph, N., DasSarma, N., Henighan, T., ... & Olah, C. (2022). In-context learning and induction heads. arXiv preprint arXiv:2209.11895.
10. Belinkov, Y. (2022). Probing classifiers: Promises, shortcomings, and advances. Computational Linguistics, 48(1), 207-219.
11. Wang, H., Minervini, P., & Ponti, E. M. (2024). Probing the emergence of cross-lingual alignment during LLM training. arXiv preprint arXiv:2406.13229.
12. Meng, K., Bau, D., Andonian, A., & Belinkov, Y. (2022). Locating and editing factual associations in gpt. Advances in neural information processing systems, 35, 17359-17372.
13. Nanda, N. (2023). Attribution patching: Activation patching at industrial scale. URL: https://www. neelnanda. io/mechanistic-interpretability/attribution-patching.
14. Jha, A., K. Aicher, J., R. Gazzara, M., Singh, D., & Barash, Y. (2020). Enhanced integrated gradients: improving interpretability of deep learning models using splicing codes as a case study. Genome biology, 21(1), 149.
15. Elhage, N., Nanda, N., Olsson, C., Henighan, T., Joseph, N., Mann, B., ... & Olah, C. (2021). A mathematical framework for transformer circuits. Transformer Circuits Thread, 1(1), 12.
16. Berkowitz, J., Weissenbacher, D., Srinivasan, A., Friedrich, N. A., Cortina, J. M. A., Kivelson, S., ... & Tatonetti, N. P. (2025, June). Probing Large Language Model Hidden States for Adverse Drug Reaction Knowledge. In International Conference on Artificial Intelligence in Medicine (pp. 55-64). Cham: Springer Nature Switzerland.
17. Lepori, M. A., Serre, T., & Pavlick, E. (2023). Uncovering intermediate variables in transformers using circuit probing. arXiv preprint arXiv:2311.04354.
18. Ou, Y., Yao, Y., Zhang, N., Jin, H., Sun, J., Deng, S., ... & Chen, H. (2025). How do llms acquire new knowledge? a knowledge circuits perspective on continual pre-training. arXiv preprint arXiv:2502.11196.
19. Wei, J., Tay, Y., Bommasani, R., Raffel, C., Zoph, B., Borgeaud, S., ... & Fedus, W. (2022). Emergent abilities of large language models. arXiv preprint arXiv:2206.07682.
20. Berti, L., Giorgi, F., & Kasneci, G. (2025). Emergent abilities in large language models: A survey. arXiv preprint arXiv:2503.05788.
21. Olsson, C., Elhage, N., Nanda, N., Joseph, N., DasSarma, N., Henighan, T., ... & Olah, C. (2022). In-context learning and induction heads. arXiv preprint arXiv:2209.11895.
22. Woodside, T. (2024). Emergent Abilities in Large Language Models: An Explainer. CSET Center for Security and Emerging Technology.
23. Du, Z., Zeng, A., Dong, Y., & Tang, J. (2024). Understanding emergent abilities of language models from the loss perspective. Advances in neural information processing systems, 37, 53138-53167.

24. Rai, D., Zhou, Y., Feng, S., Saparov, A., & Yao, Z. (2024). A practical review of mechanistic interpretability for transformer-based language models. arXiv preprint arXiv:2407.02646.
25. Power, A., Burda, Y., Edwards, H., Babuschkin, I., & Misra, V. (2022). Grokking: Generalization beyond overfitting on small algorithmic datasets. arXiv preprint arXiv:2201.02177.
26. Nanda, N., Chan, L., Lieberum, T., Smith, J., & Steinhardt, J. (2023). Progress measures for grokking via mechanistic interpretability. arXiv preprint arXiv:2301.05217.
27. Sun, Q., Pickett, M., Nain, A. K., & Jones, L. (2025, April). Transformer layers as painters. In Proceedings of the AAAI Conference on Artificial Intelligence (Vol. 39, No. 24, pp. 25219-25227).
28. Duan, X., Yao, Z., Zhang, Y., Wang, S., & Cai, Z. G. (2025). How Syntax Specialization Emerges in Language Models. arXiv preprint arXiv:2505.19548.
29. Xu, Y., Wang, Y., Huang, H., & Wang, H. (2024). Tracking the feature dynamics in llm training: A mechanistic study. arXiv preprint arXiv:2412.17626.
30. Skean, O., Arefin, M. R., Zhao, D., Patel, N., Naghiyev, J., LeCun, Y., & Shwartz-Ziv, R. (2025). Layer by layer: Uncovering hidden representations in language models. arXiv preprint arXiv:2502.02013.
31. Zhang, F., & Nanda, N. (2023). Towards best practices of activation patching in language models: Metrics and methods. arXiv preprint arXiv:2309.16042.
32. Heimersheim, S., & Nanda, N. (2024). How to use and interpret activation patching. arXiv preprint arXiv:2404.15255.
33. Achtibat, R., Hatefi, S. M. V., Dreyer, M., Jain, A., Wiegand, T., Lapuschkin, S., & Samek, W. (2024). Attnlrp: attention-aware layer-wise relevance propagation for transformers. arXiv preprint arXiv:2402.05602.
34. Wei, J. (2022, November 14). 137 emergent abilities of large language models [Blog post]. Jason Wei's Blog. Retrieved July 3, 2025, from https://www.jasonwei.net/blog/emergence
35. Chang, T. A., Tu, Z., & Bergen, B. K. (2024). Characterizing learning curves during language model pre-training: Learning, forgetting, and stability. Transactions of the Association for Computational Linguistics, 12, 1346-1362.
36. Nakagi, Y., Tada, K., Yoshino, S., Nishimoto, S., & Takagi, Y. (2025). Triple Phase Transitions: Understanding the Learning Dynamics of Large Language Models from a Neuroscience Perspective. arXiv preprint arXiv:2502.20779.
37. Ruben, R. J. (1997). A time frame of critical/sensitive periods of language development. Acta oto-laryngologica, 117(2), 202-205.
38. Jason Wei & Rishi Bommasani. (2022, September 13). Examining emergent abilities in large language models [News article]. Stanford HAI. Retrieved July 3, 2025, from https://hai.stanford.edu/news/examining-emergent-abilities-large-language-models
39. Kaplan, J., McCandlish, S., Henighan, T., Brown, T. B., Chess, B., Child, R., ... & Amodei, D. (2020). Scaling laws for neural language models. arXiv preprint arXiv:2001.08361.
40. Bahri, Y., Dyer, E., Kaplan, J., Lee, J., & Sharma, U. (2024). Explaining neural scaling laws.

Proceedings of the National Academy of Sciences, 121(27), e2311878121.
41. Gao, L., la Tour, T. D., Tillman, H., Goh, G., Troll, R., Radford, A., ... & Wu, J. (2024). Scaling and evaluating sparse autoencoders. arXiv preprint arXiv:2406.04093.
42. Ameisen, E., Lindsey, J., Pearce, A., Gurnee, W., Turner, N. L., Chen, B., Citro, C., Abrahams, D., Carter, S., Hosmer, B., Marcus, J., Sklar, M., Templeton, A., Bricken, T., McDougall, C., Cunningham, H., Henighan, T., Jermyn, A., Jones, A., Persic, A., Qi, Z., Thompson, T. B., Zimmerman, S., Rivoire, K., Conerly, T., Olah, C., & Batson, J. (2025, March 27). On the biology of a large language model [Web page]. Transformer Circuits Thread. Retrieved July 3, 2025, from https://transformer-circuits.pub/2025/attribution-graphs/biology.html
43. Xu, Y., Wang, Y., Huang, H., & Wang, H. (2024). Tracking the feature dynamics in llm training: A mechanistic study. arXiv preprint arXiv:2412.17626.
44. Clauw, K., Stramaglia, S., & Marinazzo, D. (2024). Information-theoretic progress measures reveal grokking is an emergent phase transition. arXiv preprint arXiv:2408.08944.
45. Milička, J., Marklová, A., VanSlambrouck, K., Pospíšilová, E., Šimsová, J., Harvan, S., & Drobil, O. (2024). Large language models are able to downplay their cognitive abilities to fit the persona they simulate. Plos one, 19(3), e0298522.
46. Evanson, L., Bulteau, C., Chipaux, M., Dorfmüller, G., Ferrand-Sorbets, S., Raffo, E., Rosenberg, S., Bourdillon, P., & King, J. R. (2025, May 14). Emergence of language in the developing brain. ArXiv. https://ai.meta.com/research/publications/emergence-of-language-in-the-developing-brain/
47. Hofmann, V., Weissweiler, L., Mortensen, D. R., Schütze, H., & Pierrehumbert, J. B. (2025). Derivational morphology reveals analogical generalization in large language models. Proceedings of the National Academy of Sciences, 122(19), e2423232122.
48. Holyoak, K. J. (2025, March). Analogy and the roots of creative intelligence. The MIT Press Reader. https://thereader.mitpress.mit.edu/analogy-and-the-roots-of-creative-intelligence/
49. Inani, K., Kabra, K., Marupudi, V., & Varma, S. (2025). Modeling Understanding of Story-Based Analogies Using Large Language Models. arXiv preprint arXiv:2507.10957.
50. Paulo, G., & Belrose, N. (2025). Sparse autoencoders trained on the same data learn different features. arXiv preprint arXiv:2501.16615.
51. Sharkey, L., Chughtai, B., Batson, J., Lindsey, J., Wu, J., Bushnaq, L., ... & McGrath, T. (2025). Open problems in mechanistic interpretability. arXiv preprint arXiv:2501.16496.